\documentclass{article}
\usepackage{ijcai21}

\usepackage{times}

\usepackage{soul}
\usepackage{url}
\usepackage[hidelinks]{hyperref}
\usepackage[utf8]{inputenc}
\usepackage[small]{caption}
\usepackage{graphicx}
\usepackage{amsmath}
\usepackage{amssymb}
\usepackage{booktabs}
\usepackage[lined,boxed,commentsnumbered,linesnumbered,ruled]{algorithm2e}
\usepackage{color}
\usepackage{xcolor}

\usepackage{enumitem}
\setlist{leftmargin=*}
\usepackage{xspace}
\newcommand{\method}{\texttt{SafeDrug}\xspace }

\title{\method: Dual Molecular Graph Encoders for Recommending Effective and Safe Drug Combinations}

\author{
Chaoqi Yang$^1$\and
Cao Xiao$^2$\and
Fenglong Ma$^{3}$\and
Lucas Glass$^{2}$\And
Jimeng Sun$^{1*}$\\
\affiliations
$^1$Department of Computer Science, University of Illinois Urbana-Champaign \\ $^2$Analytics Center of Excellence, IQVIA \\
$^3$College of Information Sciences and Technology, Pennsylvania State University \\
\emails
$^1$\{chaoqiy2, jimeng\}@illinois.edu,
$^2$\{cao.xiao, Lucas.Glass\}@iqvia.com,
$^3$fenglong@psu.edu
}

\begin{document}
\maketitle
\begin{abstract}
Medication recommendation is an essential task of AI for healthcare.
Existing works focused on recommending drug combinations for patients with complex health conditions solely based on their electronic health records. Thus, they have the following limitations: (1) some important data such as {\em drug molecule structures} have not been utilized in the recommendation process. (2) drug-drug interactions (DDI) are modeled implicitly, which can lead to sub-optimal results. To address these limitations, we propose a DDI-controllable drug recommendation model named \method to leverage drugs' molecule structures and model DDIs explicitly.
\method is equipped with a global message passing neural network (MPNN) module and a local bipartite learning module to fully encode the connectivity and functionality of
drug molecules. \method also has a  controllable loss function to control DDI level in the recommended drug combinations effectively.  On a benchmark dataset, our \method is relatively shown to reduce DDI by $19.43$\% and improves $2.88$\% on Jaccard similarity between recommended and actually prescribed drug combinations over previous approaches. Moreover, \method also requires much fewer parameters than previous deep learning 
based approaches, leading to faster training by about 14\% and around $2\times$ speed-up in inference. 
\end{abstract}

\section{Introduction} 

Today abundant health data such as longitudinal electronic health records (EHR) and massive biomedical data available on the web enable researchers and doctors to build better predictive models for
clinical decision making \cite{choi2016doctor,xiao2018opportunities}. Among others, recommending effective and safe medication combinations is an important task, especially for helping patients with complex health conditions~\cite{shang2019gamenet,shang2019pre}. The primary goal of recommending medication combination is to customize a safe combination of drugs for a particular patient based on the patient's health conditions. 
Earlier medication recommendation models are \textit{instance-based}, which are only based on the current hospital visit ~\cite{zhang2017leap,wang2017safe}.
As a result, a patient with newly diagnosed hypertension will likely be treated the same as another patient who has suffered chronic uncontrolled hypertension.  Such a limitation affects the safety and utility of the recommendations. To overcome this issue, \textit{Longitudinal} methods such as ~\cite{le2018dual,shang2019gamenet} are proposed. They leverage the temporal dependencies within longitudinal patient history to provide a more personalized recommendation. However, they still suffer from the following limitations.

\begin{itemize}
    \item {\bf Inadequate Medication Encoding.}  Existing works \cite{zhang2017leap,shang2019gamenet}  often represent medications using one-hot encoding. Each drug is considered a (binary) unit, ignoring that drugs in their meaningful molecular graph representation encode important drug properties, such as efficacy and safety profiles. Also, molecule substructures  are correlated with functionalities. Such knowledge can be useful for improving accuracy and safety in medication recommendations.

\item {\bf Implicit and Non-controllable DDI Modeling.} Some existing works model drug-drug interactions (DDI) via soft or indirect constraints, like  knowledge graphs (KGs) \cite{wang2017safe,mao2019medgcn} and reinforcement post-processing \cite{zhang2017leap}. These implicit handling of DDIs results in non-controllable rates in the final recommendation or sub-optimal recommendation accuracy.
\end{itemize}

To address these, we propose a DDI-controllable drug recommendation model, named \method, to leverage drugs' molecule structures and explicitly model DDIs. 
We argue it is beneficial to incorporate molecule structures in medication recommendation. 
Our \method has the following contributions:
\begin{itemize}[leftmargin=*]
\item \textbf{Dual Molecular Encoders to Capture Global and Local Molecule Patterns.} \method model firstly learns {\em patient representation}, which is fed into dual molecule encoders to capture the global pharmacological properties~\cite{brown1868connection} and the local sub-structural patterns of a drug~\cite{huang2020caster}. 
Globally, a message-passing neural network (MPNN) encoder is constructed to pass molecular information messages for the whole drug layer by layer. Drug connectivity information is well captured in multiple hops. Locally, a bipartite encoder segments drug molecules into substructures, each possibly associated with small functional groups.
In this work, the substructure representation is fed into an effective masked neural network. 
The final output of the model is obtained by element-wise integration of the global and local encoded embeddings.
\item \textbf{DDI Controllable Loss Function.} Inspired by  proportional-integral-derivative (PID) controller \cite{aastrom1995pid}, we design a new technique to adaptively combine supervised loss and unsupervised DDI constraints with considerable flexibility. During the training, the negative DDI signal would be emphasized and backpropagated if the DDI rate of individual samples is above a certain threshold/target. In the experiment, the adaptive gradient descending can balance the model accuracy and final DDI level. With a preset target, \method model can provide reliable drug combinations to meet different levels of DDI requirements.
\item \textbf{Comprehensive Evaluations.} We follow previous works~\cite{zhang2017leap,shang2019gamenet,le2018dual} and evaluate \method on a benchmark medication recommendation  dataset extracted from the
\textit{MIMIC-III} database \cite{johnson2016mimic}. \method outperforms the best baselines relatively by 19.43\% improvement in DDI reduction, 2.88\% in Jaccard similarity, and 2.14\% in F1 measure. Besides, \method requires much fewer parameters than previous deep learning based drug recommendation approaches with 14\% reduction in training time and around $2\times$ speed-up during inference. 
\end{itemize}

Data processing files and codes are released in Github~\footnote{https://github.com/ycq091044/SafeDrug}. A full version of the paper can be found in arXiv~\footnote{https://arxiv.org/abs/2105.02711}.


\section{Related Works}



\subsection{Medication Recommendation} 
Existing medication recommendation algorithms can be categorized into instance-based and longitudinal recommendation methods. Instance-based approaches focus on the patient's current health status. For example, LEAP \cite{zhang2017leap} extracts feature information from the current encounter and adopts a multi-instance multi-label drug recommendation setting. Longitudinal approaches are proposed to leverage the temporal dependencies within the clinical history, see ~\cite{choi2016doctor,xiao2018opportunities}. Among them, RETAIN \cite{choi2016retain} utilizes a two-level RNN with reverse time attention to model the longitudinal information. 
GAMENet \cite{shang2019gamenet} adopts memory neural networks and stores historical drug information as references for further prediction. These methods either model the final drug combination as multi-label binary classification \cite{choi2016retain,shang2019gamenet,shang2019pre} or by sequential decision making \cite{zhang2017leap}. Despite their initial success, these existing works still have the following limitations.
Other important data such as {\em drug molecule structures} can help augment drug recommendation but have not been utilized by existing models. Also, existing works model DDI inadequately, either via soft or indirect constraints. 

This paper proposes a DDI-controllable drug recommendation model that leverages drugs' molecule structures and models DDIs effectively.

\subsection{Molecule Representation}
Molecule graph representation remains an important topic due to the association between molecule structure and properties, e.g., efficacy and safety.  
Early days,  molecular
descriptors~\cite{mauri2006dragon} and
drug fingerprint~\cite{rogers2010extended,duvenaud2015convolutional} were commonly used to represent drug molecules.
Deep learning models were developed in recent years to learn molecule representation and model molecule substructures (a set of connected atoms). For example, \cite{huang2020caster} developed a
sequential pattern mining method on SMILES strings and modeled a pair of drugs as a set of their substring representation. Also, \cite{huang2020deeppurpose}
proposed to directly model molecule graphs using graph-based neural network models.

In this paper, inspired by \cite{huang2020deeppurpose}, \method is proposed to capture both the molecular global and local information during the recommendation.

\section{Problem Formulation} \label{sec:notation}

\noindent\textbf{Electrical Health Records (EHR).}
Patient EHR data
captures comprehensive medical histories of patients in the format of  longitudinal vectors of medical codes (e.g., diagnoses, procedures, and drugs).
Formally, EHR for patient $j$ can be represented as a sequence $\mathbf{X}_j=[\mathbf{x}_j^{(1)},\mathbf{x}_j^{(2)},\dots,\mathbf{x}_j^{(V_j)}]$, where $V_j$ is the total number of
 visits for patient $j$. The $i$-th entry of $\mathbf{X}_j$ documents the $i$-th visit of patient $j$, specified by $\mathbf{x}_j^{(i)} = [\mathbf{d}_j^{(i)}, \mathbf{p}_j^{(i)}, \mathbf{m}_j^{(i)}]$, where $\mathbf{d}_j^{(i)}\in\{0,1\}^{|\mathcal{D}|}, \mathbf{p}_j^{(i)}\in\{0,1\}^{|\mathcal{P}|},\mathbf{m}_j^{(i)}\in\{0,1\}^{|\mathcal{M}|}$, are multi-hot diagnosis,
procedure and medication vectors, respectively. $\mathcal{D}, \mathcal{P}, \mathcal{M}$
are the corresponding element sets, and $|\cdot|$
denotes the cardinality. 

\smallskip
\noindent\textbf{Safe Drug Combination Recommendation.}
We focus on drug recommendation while reducing/controlling the drug-drug interaction (DDI) rates in the prediction.
We use a symmetric binary adjacency matrix $\mathbf{D}\in
\{0,1\}^{|\mathcal{M}|\times|\mathcal{M}|}$ to denote DDI relation.
$\mathbf{D}_{ij}=1$ means that the
interaction between drug $i$ and $j$ has been reported while $\mathbf{D}_{ij}=0$
indicates a safe co-prescription. The paper aims at learning a drug recommendation function, $f(\cdot)$, such that for each visit, e.g., the $t$-th visit,
given the patient's up-to-now diagnosis and procedure sequences $[\mathbf{d}^{(1)}, \mathbf{d}^{(2)},
\dots, \mathbf{d}^{(t)}]$ and $[\mathbf{p}^{(1)}, \mathbf{p}^{(2)},
\dots, \mathbf{p}^{(t)}]$, along with the DDI matrix $\mathbf{D}$, $f(\cdot)$ can generate a drug recommendation,
\begin{equation}
    \hat{\mathbf{m}}^{(t)} = f~([\mathbf{d}^{(i)}]_{i=1}^t, [\mathbf{p}^{(i)}]_{i=1}^t)\in
\{0,1\}^{|\mathcal{M}|}.
\end{equation} The objective function consists of two parts: (i) extracting real drug combinations, $\mathbf{m}^{(t)}$, as supervision to penalize $\hat{\mathbf{m}}^{(t)}$; (ii) using matrix $\mathbf{D}$ to derive unsupervised DDI constraints on $\hat{\mathbf{m}}^{(t)}$.



\section{The \method Model} \label{sec:method}

\begin{figure*}[htbp!] \centering
	\includegraphics[width=0.87\textwidth]{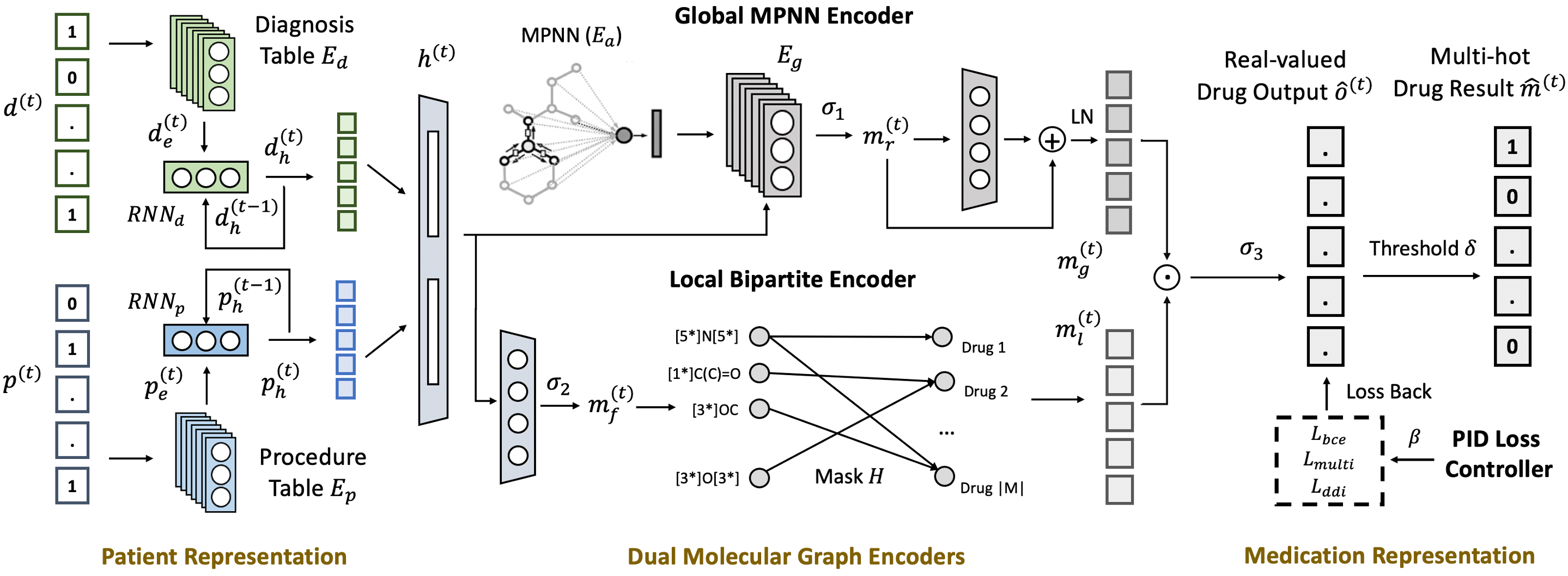} \vspace{-2mm}
	\caption{The \method
		Model. We first encode diagnosis and procedure sequences by {RNN}s to generate a patient health representation, $\mathbf{h}^{(t)}$; this representation then passes through dual molecular graph encoders for global and local molecular structural embeddings, $\mathbf{m}_g^{(t)}$ and $\mathbf{m}_l^{(t)}$; two embedding vectors are finally combined and thresholded to obtain the recommended drug combination, $\hat{\mathbf{m}}^{(t)}$.
		} \label{fig:framework} \end{figure*}

As illustrated in Figure~\ref{fig:framework}, our \method model comprises of four components: (1)  a {\bf longitudinal patient representation module} that learns patient representation from their EHR data; (2)  a {\bf global message passing neural network (MPNN) encoder} that takes patient representation as input and outputs a global drug vector with each entry quantifying the similarity between the 
patient representation and each of the drug representations; (3) in parallel, a {\bf bipartite encoder} also takes the same patient representation and outputs a local drug vector, which encodes drugs' molecular substructural functionalities; (4) finally, the global and local drug representation vectors are element-wise combined in the {\bf  medication representation module}, where the final drug output is obtained from element-wise thresholding.

\vspace{-1mm}
\subsection{Longitudinal Patient Representation} From longitudinal EHR data, patient health can be encoded by their diagnosis and procedure information.

\smallskip
\noindent
\textbf{Diagnosis Embedding.} To fully utilize rich diagnosis information, we design an embedding table,
$\mathbf{E}_d\in \mathbb{R}^{|\mathcal{D}|\times dim}$, where each row stores the embedding vector
for a particular diagnosis and $dim$ is
the dimension of the embedding space. Given a multi-hot diagnosis vector $\mathbf{d}^{(t)}\in
\{0,1\}^{|\mathcal{D}|}$, we project the corresponding diagnosis codes into the embedding space by vector-matrix dot product, which essentially takes the summation of each diagnosis embedding, \begin{equation}\label{eq:diag-emb} \mathbf{d}^{(t)}_e = \mathbf{d}^{(t)} \mathbf{E}_d.
\end{equation}
During the training, $\mathbf{E}_d$ is learnable and shared between each visit and every patient. 

\smallskip
\noindent
\textbf{Procedure Embedding.} Similarly, a shared procedure table
$\mathbf{E}_p\in \mathbb{R}^{|\mathcal{P}|\times dim}$ is also designed to encode the
associated procedure vector $\mathbf{p}^{(t)}$ (also a multi-hot vector), \begin{equation} \label{eq:prod-emb}
\mathbf{p}^{(t)}_e = \mathbf{p}^{(t)} \mathbf{E}_p. \end{equation}
Two {\em embedding vectors}, $\mathbf{d}^{(t)}_e,\mathbf{p}^{(t)}_e\in\mathbb{R}^{dim}$, encode the patient's current health condition. In fact, a health snapshot might be insufficient for treatment decisions. For example, it is
inappropriate to prescribe the same medicine for one patient with a newly
diagnosed hypertension and another patient who has chronic hypertension for more
than two years. 
Here we use RNN models to model  dynamic patient history. Specifically, we use two separate {RNN}s to obtain the {\em hidden diagnosis and procedure vectors}, $\mathbf{d}^{(t)}_h$ and $\mathbf{p}^{(t)}_h$, in case that
one of the sequences might be inaccessible in practice,
{\begin{align} \mathbf{d}^{(t)}_h &=
\mbox{RNN}_d~(\mathbf{d}^{(t)}_e,\mathbf{d}^{(t-1)}_h)
= \mbox{RNN}_d~(\mathbf{d}^{(t)}_e,\dots,\mathbf{d}^{(1)}_e), \label{eq:hidden-diag}\\
\mathbf{p}^{(t)}_h &=
\mbox{RNN}_p~(\mathbf{p}^{(t)}_e,\mathbf{p}^{(t-1)}_h)
= \mbox{RNN}_p~(\mathbf{p}^{(t)}_e,\dots,\mathbf{p}^{(1)}_e), \label{eq:hidden-prod}
\end{align} }
where $\mathbf{d}^{(t)}_h,\mathbf{p}^{(t)}_h\in\mathbb{R}^{dim}$. And $\mathbf{d}^{(0)}_h$, $\mathbf{p}^{(0)}_h$ are all-zero vectors.

\smallskip
\noindent
\textbf{Patient Representation.} 
We then concatenate the diagnosis embedding $\mathbf{d}^{(t)}_h$ and  procedure embedding $\mathbf{p}^{(t)}_h$  into a more compact {\em patient representation} $\mathbf{h}^{(t)}$. We follow a common and effective
approach  to first \textit{concatenate} two vectors as a double-long vector,
and then apply a feed-forward neural network, $\mbox{NN}_1(\cdot): \mathbb{R}^{2dim}
\mapsto \mathbb{R}^{dim}$,
($\mathbf{W}_1$ is the parameter, and $\#$ is the concatenation operation), \begin{equation}\label{eq:patient_rep} \mathbf{h}^{(t)}
= \mbox{NN}_1([\mathbf{d}^{(t)}_h ~ \# ~\mathbf{p}^{(t)}_h]; \mathbf{W}_1). \end{equation}

To this end, we obtain an overall patient representation
$\mathbf{h}^{(t)}\in\mathbb{R}^{dim}$, from which, we next generate a safe drug recommendation by comprehensively modeling the drug molecule databases. 

\vspace{-1mm}
\subsection{Dual Molecular Graph Encoders} 

To investigate drug properties and their dependencies, we exploit the
global molecule connectivity features and the local molecule functionalities. In the following sections, we introduce two new molecular graph encoders, in parallel, to comprehensively encode
the drug molecules.

\vspace{-1mm}
\subsection*{(I) Global MPNN Encoder} 
We encode drug molecule data using a message passing neural network (MPNN) operator with learnable fingerprints, aiming at convolving and pooling atom information across single molecule graphs into vector representations.

To begin with, we collect a set of all appeared atoms, $\mathcal{C}=\{a_i\}$, and design a learnable atom
embedding table, $\mathbf{E}_a\in\mathbb{R}^{|\mathcal{C}|\times dim}$, where each row
looks up the initial embedding/fingerprint for a particular atom. Our design of the MPNN model is very different from previous works \cite{coley2017convolutional,yang2019analyzing}, which mostly use atom descriptors, e.g., atom mass and chirality, as the initial features. However, for molecule structure modeling, atom-atom connectivity is more important than individual atoms.

Given a
drug molecule graph (atoms $a_0, \dots, a_n$ as the vertices and atom-atom bonds as
edges) with adjacency matrix $\mathbf{A}$ and the initial atom fingerprints from $\mathbf{E}_a$,
i.e., $\{\mathbf{y}_0^{(0)},\mathbf{y}_1^{(0)},\dots, \mathbf{y}_n^{(0)} \}$, we formulate the layer-wise message passing over the graph as,
\begin{align}
\mathbf{z}_i^{(l+1)} &= \sum_{j:~\mathbf{A}_{ij}=1}\mbox{MESSAGE}_l~(\mathbf{y}_i^{(l)}, \mathbf{y}_j^{(l)}; \mathbf{W}^{(l)}),
\label{eq:z_i}\\ 
\mathbf{y}_i^{(l+1)} &= \mbox{UPDATE}_l~(\mathbf{y}_i^{(l)}, \mathbf{z}_i^{(l+1)}),~i=0,\dots,n \label{eq:y_i}\end{align} 
where $l$ is the layer index, $\mathbf{z}_i^{(l+1)}$ is the encoded messages from the neighbors of atom $i$ at the $l$-th iteration, $\mathbf{y}_i^{(l+1)}$ is the hidden state of atom $i$. During message passing, $\mathbf{z}_i^{(l+1)}$ and $\mathbf{y}_i^{(l+1)}$ with each vertex atom $i$ are updated using {\em message function}, $\mbox{MESSAGE}_l(\cdot)$, and vertex {\em update function}, $\mbox{UPDATE}_l(\cdot)$, and $\mathbf{W}^{(l)}$ is the layer-wise parameter matrix. After applying message passing for $L$ layers, the global representation of a drug molecule is pooled
by a {\em readout function}, which calculates the average of all atom fingerprints, \begin{equation} \label{eq:y} \mathbf{y} = \mbox{READOUT}~(\{\mathbf{y}_i^{(L)}\mid i=0,\dots,n\}). \vspace{1mm}\end{equation} 
We apply the same MPNN encoder with shared parameters for each single drug moledule (there are $|\mathcal{M}|$
different drugs in total). We collect the MPNN embeddings of all drug molecules into a drug memory  $\mathbf{E}_g\in\mathbb{R}^{|\mathcal{M}|\times dim}$,
where each row is the embedding of a drug.

\smallskip
\noindent
\textbf{Patient-to-drug Matching.} When treating patient
representation $\mathbf{h}^{(t)}$ as a query, we aim at searching the most {\em relevant} drugs
from the memory $\mathbf{E}_g$. The health-to-drug matching score is calculated by
the dot product of the MPNN representation and the patient representation $\mathbf{h}^{(t)}$. For all drugs, we operate each row of $\mathbf{E}_g$ with $\mathbf{h}^{(t)}$, followed by a sigmoid function $\sigma_1(\cdot)$,
\begin{align} \label{eq:relevant}\mathbf{m}_r^{(t)} &= \sigma_1~(\mathbf{E}_g \mathbf{h}^{(t)} ).\end{align}
Every element of $\mathbf{m}_r^{(t)}$ stores a matching score for one drug. In light of the recent successful post-LN architecture \cite{wang2019learning} in Transformer \cite{vaswani2017attention}, the matching score is further parameterized by a feed-forward neural network, $\mbox{NN}_2(\cdot): \mathbb{R}^{dim}
\mapsto \mathbb{R}^{dim}$, followed by layer-normalization (LN) step with residual connection,
\begin{align}\label{eq:global} \mathbf{m}_g^{(t)} &= \mbox{LN}~(\mathbf{m}_r^{(t)}+\mbox{NN}_2~(\mathbf{m}_r^{(t)};\mathbf{W}_2)).\end{align}

\subsection*{(II) Local Bipartite Encoder} 

Molecules with similar
structures shall be mapped into nearby space using the MPNN encoder.  However, 
a pair of drugs that have significant overlap on structure domain might behave
differently in terms of DDI interaction or other functional activities
\cite{ryu2018deep}. 
In fact, a drug's functionality is better reflected by molecule
substructures, which are defined as a set of connected atoms that may have associated functional groups. This section focuses on the substructure composition of molecules and builds a  substructure-to-drug bipartite architecture for capturing drug local functionality and their dependencies.

\smallskip
\noindent
\textbf{Molecule Segmentation.} 
To capture the relations between drug molecules and their substructures, we first adopt the well-known molecule segmentation method, breaking retrosynthetically interesting chemical substructures (BRICS) \cite{degen2008art}, which preserves the most critical drug functional groups and more crucial bonds.  \method can easily support other segmentation methods such as RECAP, ECFP.
  After the segmentation, each molecule is decomposed
into a set of chemical substructures. 
In this setting, a pair of drugs could intersect on a mutual set of substructures. We denote the overall substructure set as $\mathcal{S}$. Based on the many-to-many relations, it is ready to construct a bipartite architecture, represented by a {\it mask matrix} $\mathbf{H}\in \{0,1\}^{|\mathcal{S}|\times |\mathcal{M}|} $: $\mathbf{H}_{ij}=1$ indicates that drug $j$ contains  substructure $i$.

\smallskip
\noindent
\textbf{Bipartite Learning.} 
We hope that $\mathbf{H}$ can help to derive a drug representation on its functionality information, given the input, $\mathbf{h}^{(t)}$. To do so, we first apply a feed-forward neural network,
$\mbox{NN}_3(\cdot): \mathbb{R}^{dim}
\mapsto \mathbb{R}^{|\mathcal{S}|}$, for dimensionality transformation, followed by a sigmoid function, $\sigma_2(\cdot)$. In fact, $\mbox{NN}_3(\cdot)$ transforms the patient representation $\mathbf{h}^{(t)}$ into a local functional vector,
\begin{equation} \label{eq:functional}
    \mathbf{m}_f^{(t)} = \sigma_2~(\mbox{NN}_3~(\mathbf{h}^{(t)}; ~\mathbf{W}_3)).
\end{equation}

Each entry of $\mathbf{m}_f^{(t)}$ quantifies the significance of the corresponding functionality. Therefore, $\mathbf{m}_f^{(t)}$ can be viewed as an  anti-disease functionality combination for the current patient. Next, we are motivated to generate the recommendation of drugs
that can cover all these anti-disease functionalities while considering to prevent DDIs.

We therefore perform network pruning and design a masked 1-layer neural network, $\mbox{NN}_4(\cdot): \mathbb{R}^{|\mathcal{S}|}
\mapsto \mathbb{R}^{|\mathcal{M}|}$, where the parameter matrix is masked by the bipartite architecture $\mathbf{H}$, essentially through matrix element-wise product $\odot$. During the training process, $\mbox{NN}_4(\cdot)$ will learn to map the functionality vector $\mathbf{m}_f^{(t)}$ into the corresponding local drug representation,
\begin{align} \label{eq:mask}
\mathbf{m}_l^{(t)} =
\mbox{NN}_4~(\mathbf{m}_f^{(t)}; ~\mathbf{W}_4\odot \mathbf{H}). 
\end{align}

We show that the mask $\mathbf{H}$ would enable the model to (i) have far fewer parameters (due to the sparsity of $\mathbf{H}$); (ii) prevent DDIs by avoiding a co-prescription of interacted drugs (we demonstrate it in Appendix).



\subsection{Medication Representation} 

The final medication representation is
given by treating the global drug matching vector as attention signals to further adjust
the local drug functionality representation. We finally use a sigmoid function $\sigma_3(\cdot)$ to scale the output,
\begin{equation}\label{eq:final} \hat{\mathbf{o}}^{(t)} = \sigma_3~(\mathbf{m}_g^{(t)} \odot \mathbf{m}_l^{(t)}), \end{equation}
where $\odot$ represents the element-wise product. By a threshold value
$\delta$, we can achieve the recommended drug combination, i.e., a multi-hot vector
$\hat{\mathbf{m}}^{(t)}$, by picking those entries where the value is greater than $\delta$.

\subsection{Model Training and Inference} 

\method is trained end-to-end. We simultaneously learn  $\mathbf{E}_d$, $\mathbf{E}_p$, parameters in $\mbox{RNN}_d$
and $\mbox{RNN}_p$, network parameters $\mathbf{W}_1$,  $\mathbf{E}_a$,
layer-wise parameters in MPNN $\{\mathbf{W}^{(i)}\}$, parameters in layer normalization (LN), and $\mathbf{W}_2$, $\mathbf{W}_3$ and $\mathbf{W}_4$.

\smallskip
\noindent
\textbf{Multiple Loss.} In this paper, the recommendation task is
formulated as multi-label binary classification. Suppose
$|\mathcal{M}|$ is the total number of drugs. We use $\mathbf{m}^{(t)}$ to denote the target
multi-hot drug recommendation, 
$\hat{\mathbf{o}}^{(t)}$ for the output real-valued drug representation (opposed to the predicted multi-hot vector $\hat{\mathbf{m}}^{(t)}$ after applying threshold $\delta$). We consider the
prediction of each drug as a sub-problem and use binary cross-entropy
(BCE) as a loss, 
\begin{equation} \label{eq:bce} L_{bce} =
-\sum_{i=1}^{|\mathcal{M}|} \mathbf{m}^{(t)}_i log (\hat{\mathbf{o}}^{(t)}_i) + (1-\mathbf{m}^{(t)}_i)log (1-\hat{\mathbf{o}}^{(t)}_i),
\end{equation} where $\mathbf{m}^{(t)}_i$ and $\hat{\mathbf{o}}^{(t)}_i$ are the $i$-{th} entries. To make the
result more robust, we also adopt multi-label hinge loss to ensure
that the truth labels have at least 1 margin larger than others,
{\small \begin{equation} \label{eq:multi}
L_{multi} = \sum_{i,j:~\mathbf{m}^{(t)}_i=1,\mathbf{m}^{(t)}_j= 0}
\frac{\mbox{max}(0,1-(\hat{\mathbf{o}}^{(t)}_i-\hat{\mathbf{o}}^{(t)}_j))}{|\mathcal{M}|}. \end{equation}}

\noindent Based on the DDI adjacency matrix $\mathbf{D}$, we also design the adverse DDI loss over $\hat{\mathbf{o}}^{(t)}$ as, 
\begin{equation}  \label{eq:ddi}
L_{ddi} =
\sum_{i=1}^{|\mathcal{M}|}\sum_{j=1}^{|\mathcal{M}|} \mathbf{D}_{ij} \cdot \hat{\mathbf{o}}^{(t)}_i
\cdot \hat{\mathbf{o}}^{(t)}_j, \end{equation}
where $\cdot$ is the product between scalars. Note that the above loss functions are defined
for a single visit.  During the training, loss back propagation will be conducted at patient level by the averaged losses across all visits.

\smallskip
\noindent
\textbf{Controllable Loss Function.} A standard approach to training with
multiple loss functions is by the weighted sum of the loss measuring terms
\cite{dosovitskiy2019you}, 
\begin{equation}  \label{eq:alpha}
L = \beta\left(\alpha L_{bce} +
(1-\alpha)L_{multi}\right) + \left(1-\beta\right)L_{ddi}, \end{equation} where $\alpha$ and
$\beta$ are usually pre-defined hyperparameters. 

In this scenario, two prediction
loss, $L_{bce}$ and $L_{multi}$, are compatible, and thus we select $\alpha$
from the validation set. We observe that adverse DDIs also exist in the dataset
(doctors might mistakenly prescribe the interacted drugs). Therefore, by training with ground truth labels,
the correct learning process might conversely increase DDI. Thus,
we consider adjusting $\beta$ during the training process by a
proportional–integral–derivative  (PID) controller \cite{an2018pid} to balance the prediction
loss and DDI loss. For simplicity, we only utilize the proportional error signal as negative feedback when the DDI rate of the recommended drugs is above certain thresholds.

In safe drug recommendation, the key is to maintain a low DDI, so we set a DDI
acceptance rate $\gamma \in (0,1)$ for the loss function.
Our motivation is that if the patient-level DDI is lower than the threshold, $\gamma$, then we consider only maximizing the prediction accuracy; otherwise, $\beta$ will adjust adaptively to reduce DDI as well.

The controllable
factor $\beta$ follows the piece-wise strategy, \begin{equation}\label{eq:loss}
\beta =\left\{ \begin{array}{lr} 1, ~~& \mbox{DDI} \leq \gamma \\ \max~\{0, ~1 -
\frac{DDI - \gamma}{K_p}\}, ~~& \mbox{DDI} > \gamma \end{array} \right.
\end{equation}
where $K_p$ is a correcting factor for the proportional
signal. %

We show in the experiment that $\gamma$ can be viewed as an upper bound of the
output DDI rate. By presetting  $\gamma$, our \method 
can meet different level of DDI
requirements.

The model inference phase basically follows the same pipeline as training. We apply a threshold $\delta=0.5$ on the output drug representation in Eqn.~\eqref{eq:final} and select the drugs, corresponding to entries where the value is greater than $\delta$, as the final recommendations. We structure the overall algorithm in Appendix.

\section{Experiments} 






\vspace{-1mm}

\noindent
\textbf{Dataset and Metrics.} The experiments are carried out on \textit{MIMIC-III} \cite{johnson2016mimic}. The statistics of the post-processed data is
reported in Table~\ref{tb:mimic}. We use five efficacy metrics: DDI rate, Jaccard similarity, F1 score, PRAUC, and \# of medications, to evaluate the recommendation efficacy and three metrics: \# of parameters, training time, and inference time, to evaluate the complexity of the models. Due to space limitation, we push other details to Appendix.

\begin{table}[h!] \small
\centering
	\begin{tabular}{l|c} \toprule \textbf{Items} & \textbf{Size} \\ \midrule
		\# of visits / \# of patients & 14,995 / 6,350 \\ diag. / prod. /
		med. space size & 1,958 / 1430 / 131 \\ 
		avg. / max \#  of visits & 2.37 / 29 \\ 
		avg. / max \#  of diagnoses per visit & 10.51 / 128 \\ 
		avg. / max \#  of procedures per visit & 3.84 / 50 \\ 
		avg. / max \#  of medicines per visit & 11.44 / 65 \\ 
		total \# of DDI pairs & 448 \\
		total \# of substructures & 491 \\
\bottomrule \end{tabular}
    \vspace{-2mm}
    \caption{Data Statistics}
    \vspace{-3mm}
	\label{tb:mimic} \end{table}

\begin{table*}[!ht] 
\centering
{
	\resizebox{.93\textwidth}{!}
	{
	\begin{tabular}{l|ccccc} \toprule {\bf Model} & {\bf DDI} & {\bf Jaccard}
			& {\bf F1-score} & {\bf PRAUC} & {\bf Avg. \# of Drugs}\\
			\midrule LR & 0.0829 $\pm$ 0.0009 (0.0) & 0.4865 $\pm$ 0.0021 (0.0) & 0.6434 $\pm$ 0.0019 (0.0) & 0.7509 $\pm$ 0.0018 (3e-11) & 16.1773 $\pm$ 0.0942\\ 
			ECC  & 0.0846 $\pm$ 0.0018 (0.0) & 0.4996 $\pm$ 0.0049 (5e-10) & 0.6569 $\pm$ 0.0044 (4e-10) & 0.6844 $\pm$ 0.0038 (0.0) & 18.0722 $\pm$ 0.1914\\
				\midrule
			RETAIN  & 0.0835 $\pm$ 0.0020 (0.0) & 0.4887 $\pm$ 0.0028 (2e-15) & 0.6481 $\pm$ 0.0027 (5e-15) & 0.7556 $\pm$ 0.0033 (2e-06) & 20.4051 $\pm$ 0.2832\\ 
			LEAP   & {0.0731 $\pm$ 0.0008} (0.0) & 0.4521 $\pm$ 0.0024 (0.0) & 0.6138 $\pm$ 0.0026 (0.0) & 0.6549 $\pm$ 0.0033 (0.0) & 18.7138 $\pm$ 0.0666\\ 
			DMNC   & 0.0842 $\pm$ 0.0011 (0.0) & 0.4864 $\pm$ 0.0025 (2e-16) & 0.6529 $\pm$ 0.0030 (3e-13) & 0.7580 $\pm$ 0.0039 (2e-04) & 20.0000 $\pm$ 0.0000\\ 
			GAMENet   & 0.0864 $\pm$ 0.0006 (0.0) & {0.5067 $\pm$ 0.0025} (6e-10) & {0.6626 $\pm$ 0.0025} (4e-10) & {0.7631 $\pm$ 0.0030} (0.21) & 27.2145 $\pm$ 0.1141\\ 
			\midrule
			$\method_{L}$ & \textbf{0.0580 $\pm$ 0.0004} ($- - $) & 0.5166 $\pm$ 0.0026 (1e-03) & 0.6670 $\pm$ 0.0024 (9e-08) & 0.7632 $\pm$ 0.0026 (0.20) & 18.6663 $\pm$ 0.1417\\ 
			$\method_{G}$ & 0.0606 $\pm$ 0.0007 (7e-06) & 0.4862 $\pm$ 0.0027 (4e-16) & 0.6442 $\pm$ 0.0024 (2e-16) & 0.7423 $\pm$ 0.0029 (4e-13) & 19.0724 $\pm$ 0.0971\\
			\method  & {0.0589 $\pm$ 0.0005} & \textbf{0.5213 $\pm$ 0.0030} & \textbf{0.6768 $\pm$ 0.0027} & \textbf{0.7647 $\pm$ 0.0025} & 19.9178 $\pm$ 0.1604\\
			\bottomrule
\end{tabular}}}
\vspace{-2mm}
\caption{Performance Comparison on
		\textit{MIMIC-III} (ground truth DDI rate is
	0.0808)}\label{tb:general}

\end{table*}

\smallskip
\noindent
\textbf{Baselines.} We compare \method with the following
baselines from different perspectives: standard Logistic Regression (LR), multi-label classification approach: Ensemble Classifier Chain (ECC), \cite{read2009classifier}, RNN-based approach: RETAIN \cite{choi2016retain}, instance-based approach: LEAP \cite{zhang2017leap}, longitudinal memory-based approaches: DMNC \cite{le2018dual} and GAMENet \cite{shang2019gamenet}. \cite{shang2019pre} is not considered a baseline since it requires extra ontology data. We also compared \method with its two model variants: $\method_{L}$
is for the model with only bipartite encoder (local, L) and $\method_{G}$ for the model with only the MPNN encoder (global, G). In implementation, we use $\mathbf{m}_l^{(t)}$ and $\mathbf{m}_g^{(t)}$ as the final drug representation for them, respectively.



\smallskip
\noindent\textbf{Performance Comparison.}
From Table~\ref{tb:general},
it is easy to see the drug sequence generation models LEAP, DMNC, and ECC yield poor results among the baselines. We
, therefore, conclude that formulating the drug recommendation task as
multi-label prediction might be more straightforward and effective. LR and RETAIN do not consider the
DDI information, so they output undesirable DDI rate.  The recommendations given by GAMENet are based on historical
combinations, however, real medication records usually contain high DDI rate,
e.g., around 8\% in \textit{MIMIC-III}, so it also provides a high DDI rate. Moreover, the performance of two variants, $\method_{L}$ and $\method_{G}$, show that both of the molecular
encoders do help with the prediction. Overall, our \method also outperforms the other baselines with significantly lower
DDI rate and better accuracy.

\smallskip
\noindent
\textbf{Significance Test.} 
The above results are given by 10 rounds of bootstrapping sampling. Next, we conduct T-test on each metric
for two-tailed hypothesis. The $p$-values are shown in the parenthesis. Commonly, a $p$-value smaller than 0.05 would be considered as significant. We also use ($- - $) for cases where the result of \method is worse than the baseline. To conclude, our model can beat all compared baselines in almost every metric with $p$=0.001 significance level.

\begin{table}[h!] \small
{
	{\begin{tabular}{l|ccc} \toprule {\bf Model} & {\bf\# of Param.} & {\bf Training Time} & {\bf Inference} \\ \midrule
			RETAIN & 287,940 &
			36.35s / Epoch & 3.98s
			\\ 
			LEAP  &
			433,286 & 336.14s / Epoch & 32.31s
			\\ DMNC  &
			525,533 & 4697.40s / Epoch & 262.19s
			\\GAMENet  & 449,092 & 162.10s / Epoch & 26.85s \\ \midrule 
			$\method_{L}$  & 299,404 &
			133.43s / Epoch & 9.93s \\ 
			$\method_{G}$  & 324,577 & 129.66s / Epoch & 9.79s \\ 
			
			\method  & 325,473 & 138.77s / Epoch & 10.64s\\
			\midrule
			$\Delta$ (improve.) & $\downarrow$ 24.88\% & $\downarrow$ 14.39\% & $\downarrow$ 44.21\%\\
			\bottomrule \end{tabular}} }
			\vspace{-2mm}
			\caption{DL Model Complexity Comparison}\label{tb:comparison2}
\end{table}

\noindent\textbf{Comparison of Model Efficiency.}
Model complexity of \method and other DL baselines are evaluated in Table~\ref{tb:comparison2}. 
Since RETAIN is not specialized for drug recommendation, improvements in Table~\ref{tb:comparison2} are over LEAP, DMNC, and GAMENet. Efficiency wise, our model has relatively lower space and time complexity over LEAP, DMNC, and GAMENet. Also, it is much more efficient during inference. Leap and DMNC adopt sequential modeling and recommend drugs one by one, which is time-consuming. GAMENet stores a large memory bank, and thus it requires larger space. By comparison, we conclude that \method is
efficient and flexible for real deployment.

\smallskip
\noindent\textbf{Analysis of DDI Controllability}.
We evaluate DDI controllability and show that the average DDI rate
of our output can be controlled by threshold $\gamma$.
The ground truth DDI rate in \textit{MIMIC-III} is $0.0808$. We consider that learning-based drug recommendation approaches should optimize the final recommendations by processing EHR and drug-drug relation databases, so the output DDI rate is supposed to be lower than human experts'. In this section, we test for a series of targeted DDI $\gamma$
ranging from $0$ to $0.08$. For each $\gamma$, we
train a separate model. The experiment is conducted for 5 times. After convergence, we show the average metrics in
Table~\ref{tb:control}.

\begin{table}[t!] \small
\centering 
	\resizebox{0.4\textwidth}{!}{\begin{tabular}{c|ccccc} 
	\toprule 
	$\mathbf{\gamma}$ & {\bf DDI} & {\bf\#
	of Med.} & {\bf Jaccard}  & {\bf F1-score} & {\bf PRAUC} \\ 
	\midrule 
	0.00 & 0.0073  & 13.82 & 0.4198 & 0.5743 & 0.6902 \\ 
	0.01 & 0.0152  & 15.35 & 0.4391 & 0.5924 & 0.6788 \\ 
	0.02 & 0.0254  & 17.64 & 0.4741 & 0.6255 & 0.7196 \\ 
	0.03 & 0.0301  & 18.33 & 0.4959 & 0.6540 & 0.7492 \\ 
	0.04 & 0.0381  & 18.87 & 0.5083 & 0.6651 & 0.7556 \\ 
	0.05 & 0.0479  & 19.08 & 0.5164 & 0.6720 & 0.7604 \\ 
	0.06 & 0.0589  & 19.91 & 0.5213 & 0.6768 & 0.7647 \\ 
	0.07 & 0.0672  & 20.18 & 0.5228 & 0.6776 & 0.7634 \\ 
	0.08 & 0.0722  & 20.75 & 0.5250 & 0.6794 & 0.7691 \\
	\bottomrule \end{tabular}}
	\vspace{-2mm}
	\caption{Performance under Acceptance DDI Rate $\gamma$} 
	\vspace{-2mm}
\label{tb:control}\end{table}

From the result, the DDI
rates are controlled and upper-bounded by $\gamma$ in most
cases, which is consistent with the design. When $\gamma$ becomes larger, more drugs are allowed in
one combination, and \method is more accurate. When
$\gamma$ is too small ($<0.02$), model accuracy will drop dramatically. The parameter $\gamma$ provides a way for doctors to control the tradeoff between DDI rates and accuracy in recommendations.

\vspace{-1mm}\section{Conclusion} \vspace{-1mm} In this paper, we propose 
\method accurate and safe drug recommendation. Our model
extracts and encode rich molecule structure information to augment the drug recommendation task. We design an adaptive loss controller to provide additional flexibility. We evaluated \method using benchmark data and showed better accuracy and efficiency.


\vspace{-1mm}
\section*{Acknowledgements}

This work was in part supported by the National Science Foundation award SCH-2014438, PPoSS 2028839, IIS-1838042, the National Institute of Health award NIH R01 1R01NS107291-01 and OSF Healthcare. 

\clearpage

\bibliographystyle{named}
\bibliography{ijcai}

\clearpage

\appendix

\section{Algorithm}

We summarize the procedure of one training epoch of our model in Algorithm~\ref{algo:safedrug}. From {\em Line 1} to {\em Line 3}, we input some data sources, initialize the parameters by random $Uniform(-0.1, 0.1)$ distribution and generate a bipartite mask $\mathbf{H}$ directly from the molecule fragmentation result. From {\em Line 4} to {\em Line 8}, we select one patient from the training set and enumerate her/his clinical visits. In {\em Line 9}, we embed the longitudinal diagnosis and procedure information into a patient representation, $\mathbf{h}^{(t)}$. Note that, our algorithm would store the hidden representation, $\mathbf{d}_h^{(t)}$ and $\mathbf{p}_h^{(t)}$ for the next visit. In {\em Line 10} and {\em Line 11}, we generate the global and local drug vectors by MPNN encoder and bipartite encoder. We combine these two vectors in {\em Line 12} and compute losses and conduct gradient descent in {\em Line 14 and Line  15}.

\begin{algorithm}[h!]\scriptsize
 \SetAlgoLined
 \textbf{Input:} Training set $\mathcal{T}$, hyperparameters $\alpha, K_p$, pre-defined acceptance DDI rate $\gamma$ in Eqn.~\eqref{eq:alpha} and \eqref{eq:loss}, DDI matrix $\mathbf{D}$\;
 Initialize parameters: $\mathbf{E}_d,\mathbf{E}_p,\mathbf{E}_a,\mbox{RNN}_d, \mbox{RNN}_p, \mathbf{W}_i,\{\mathbf{W}^{(i)}\},\mbox{LN}$\;
 Obtain the bipartite mask, $\mathbf{H}$, by BRICS fragmentation \cite{degen2008art}\;
 \For{patient $j$ := 1 to $|\mathcal{T}|$}{
    Select the patient $j$'s history, $\mathbf{X}_j$ \;
    Obtain the global drug memory, $\mathbf{E}_g$, in Eqn.~\eqref{eq:z_i}, \eqref{eq:y_i} and \eqref{eq:y}\;
    \For{visit $t$ := 1 to $|\mathbf{X}_j|$}{
    Select the $t$-th visit of patient $j$, $\mathbf{x}_j^{(t)}$\;
    Generate embeddings, $\mathbf{d}_e^{(t)},\mathbf{p}_e^{(t)}$, in Eqn.~\eqref{eq:diag-emb}, \eqref{eq:prod-emb}; hidden representations, $\mathbf{d}_h^{(t)},\mathbf{p}_h^{(t)}$, in
    Eqn.~\eqref{eq:hidden-diag}, \eqref{eq:hidden-prod};
    and patient representation, $\mathbf{h}^{(t)}$, in Eqn.~\eqref{eq:patient_rep} // Patient Representation\;
    Generate matching vector, $\mathbf{m}_r^{(t)}$, and global drug vector, $\mathbf{m}_g^{(t)}$, from Eqn.~\eqref{eq:relevant} and \eqref{eq:global} // MPNN Encoder\;
    Generate functional vector, $\mathbf{m}_f^{(t)}$, and local drug vector, $\mathbf{m}_l^{(t)}$, from Eqn.~\eqref{eq:functional} and \eqref{eq:mask} // Bipartite Encoder\;
    Generate drug representation, $\hat{\mathbf{o}}_l^{(t)}$, and multi-hot drug vector, $\hat{\mathbf{m}}^{(t)}$, from Eqn.~\eqref{eq:final} // Drug Representation\;
 }
 Accumulate $L_{bce},L_{multi},L_{ddi}$ in Eqn.~\eqref{eq:bce}, \eqref{eq:multi}, \eqref{eq:ddi}\;
 }
 Choose $\beta$ and Optimizing the loss in Eqn.~\eqref{eq:alpha}, \eqref{eq:loss}\; 
 \caption{One Training Epoch of \method}
 \label{algo:safedrug}
\end{algorithm}

\section{Additional Experiment Information}

\smallskip
\noindent
\textbf{Dataset Processing.} The  \textit{MIMIC-III} \cite{johnson2016mimic}.
dataset
is publicly available with patients who stayed in intensive care unit (ICU) in the Beth Israel
Deaconess Medical Center for over 11 years. It consists of 50,206 medical
encounter records. 
We filter out the patients with only one visit. 
Diagnosis, procedure and medication information is initially documented in ``DIAGNOSES\_ICD.csv", ``PROCEDURES\_ICD.csv" and ``PRESCRIPTIONS.csv" in the original MIMIC sources. We extract them separately and then merge these three sources by patient id and visit id. After the merging, diagnosis and procedure are ICD-9 coded, and they will be transformed into multi-hot vectors before training. We extract the DDI information by Top-40 severity types from TWOSIDES \cite{tatonetti2012data}, which are reported by ATC Third Level codes. This paper then transforms the NDC drug codings to the same ATC level codes for integration, and in the implementation, we also average the drug-level MPNN embeddings and aggregate the substructures for the corresponding ATC Third level code. For molecule connectivity information, we first get the SMILES strings of the molecules from {\em drugbank.com}, then we use {\em rdkit.Chem} software to extract the molecule adjacency matrix by treating atoms as nodes and chemical bonds as edges.

\smallskip
\noindent
\textbf{Baselines Details.} We compare \method with the following
baselines. To carry out fair
comparison, we also use 64 hidden units or 64-dim embedding tables for all the
modules in the baselines. \begin{itemize} 
	\item \textbf{Logistic Regression (LR).} This is an instance-based classifier
	with $L_2$ regularization. We directly feed the concatenated multi-hot diagnosis and
	procedure vector as the feature, and medication vector as the target. The
	model is implemented by \textit{scikit-learn} with One-vs-Rest classifier, 500 maximum iteration, LBFGS algorithm as solver.

    \item
	\textbf{Ensemble Classifier Chain (ECC)} \cite{read2009classifier}. Classifier chain (CC) is a popular multi-label classification approach, which feeds previous classification results into the latter classifiers. We implement a 10-member ensemble of CCs also by \textit{scikit-learn}, where each CC consists of a dependent series of logistic regression classifiers with max iteration at 500 and LBFGS as optimizer.
	
	\item \textbf{RETAIN} \cite{choi2016retain} is an entirely temporal based
	method. It utilizes a two-level RNN with reverse time attention to model
	the longitudinal information. Following the original paper \cite{choi2016retain}, we choose two 64-dim GRUs as the implementation of two-level RNN with dropout rate at 0.5 for the output embedding. Adam is used as the optimizer with learning rate at $5\times 10^{-4}$ for RETAIN and other deep learning baselines.

	\item \textbf{LEAP} \cite{zhang2017leap} is a
	sequential decision making algorithm. It treats drug recommendation as a
	sentence generation task and recommend drugs one at a time. LEAP is also implemented by 64-dim GRU as the feature encoder. However, we choose droput rate at 0.3 between two layers, as 0.3 works better than 0.5 in the validation set. Particularly, since LEAP and DMNC are sequence-based models, we set 20 as their max drug combination size.
	
	\item \textbf{DMNC} \cite{le2018dual}. DMNC builds the drug recommendation model on differentiable neural computer (DNC). It models the diagnosis and procedure sequences by multi-view learning and uses a memory augmented neural networks (MANN). We use 64-dim embedding tables to represent medical codes, design 1-head, 1-layer, 16-cell DNC encoder using {\em dnc} package, and a 64-dim GRU decoder.

	\item \textbf{GAMENet} \cite{shang2019gamenet}. GAMENet is a recent work.
	This method also adopts memory augmented neural networks and stores
	historical drug memory records as references for future prediction. We use the same suit of hyperparameters reported from the original paper \cite{shang2019gamenet}: expected DDI rate at 0.05, initial annealing temperature at 0.85, mixture weights $\pi=[0.9, 0.1]$ and also use 64-dim embedding tables and 64-dim GRU as RNN. We find the hyperparameter works well in validation set, so we keep it for test.
\end{itemize}

\smallskip
\noindent
\textbf{Metrics Details.}
We provide the detailed definition of evaluation metrics,
\begin{itemize} [leftmargin=*]
    \item \textbf{DDI rate} (smaller is better) for patient $j$ is calculated as below, \begin{equation}
\mathrm{DDI}_j = \frac{\sum^{V_j}_{t=1}\sum_{k, l\in \{i:~\hat{\mathbf{m}}^{(t)}_{j,i}=1\}} \mathbf{1}\{\mathbf{D}_{kl}=1\}}{\sum^{V_j}_{t=1}\sum_{k, l\in \{i:~\hat{\mathbf{m}}^{(t)}_{j,i}=1\}}1}, \notag
\end{equation} 
where $V_j$ denotes the total number of visits of the $j$-th patient, $\hat{\mathbf{m}}^{(t)}_j$ represents the predicted drug combination during  the $t$-{th} visit of patient $j$, $\hat{\mathbf{m}}^{(t)}_{j,i}$ denotes the $i$-th element of $\hat{\mathbf{m}}^{(t)}_j$, $\mathbf{D}$ is the DDI matrix (defined in Section~\ref{sec:notation}) and $\mathbf{1}\{\cdot\}$ is an indicator function, which returns 1 when the expression in $\{\cdot\}$ is true, otherwise 0. 
\item \textbf{The Jaccard coefficient} (higher is better) for patient $j$ at the $t$-th visit is calculated as follows,
\begin{equation}
    \mathrm{Jaccard}_j^{(t)} = \frac{|\{i:~\mathbf{m}_{j,i}^{(t)}=1\}\cap \{i:~\hat{\mathbf{m}}_{j,i}^{(t)}=1\}|}{|\{i:~\mathbf{m}_{j,i}^{(t)}=1\}\cup \{i:~\hat{\mathbf{m}}_{j,i}^{(t)}=1\}|},
\end{equation}
where ${\mathbf{m}}^{(t)}_j$ is the ground-truth drug combination during  the $t$-{th} visit of patient $j$, and ${\mathbf{m}}^{(t)}_{j,i}$ is the $i$-th element. $|\cdot|$ denotes the cardinality, $\cap$ is set interaction and $\cup$ is set union operation. Then, the Jaccard coefficient for one patient $j$ is by taking the average over her/his visits ($V_j$ is the total number of visits of patient $j$),
\begin{equation}
    \mathrm{Jaccard}_j = \frac{1}{V_j}\sum_{t=1}^{V_j}\mathrm{Jaccard}_j^{(t)}.
\end{equation}
\item \textbf{The F1 score} (higher is better) is  the harmonic mean of precision and recall. For patient $j$ at visit $t$, the Precision, Recall, F1 are calculated by
\begin{equation}
    \mathrm{Precision}_j^{(t)} = \frac{|\{i:~\mathbf{m}_{j,i}^{(t)}=1\}\cap \{i:~\hat{\mathbf{m}}_{j,i}^{(t)}=1\}|}{ |\{i:~\hat{\mathbf{m}}_{j,i}^{(t)}=1\}|},
\end{equation}
\begin{equation}
    \mathrm{Recall}_j^{(t)} = \frac{|\{i:~\mathbf{m}_{j,i}^{(t)}=1\}\cap \{i:~\hat{\mathbf{m}}_{j,i}^{(t)}=1\}|}{ |\{i:~{\mathbf{m}}_{j,i}^{(t)}=1\}|},
\end{equation}
\begin{equation}
    \mathrm{F1}_j^{(t)} = \frac{2}{\frac{1}{\mathrm{Precision}_j^{(t)}}+\frac{1}{\mathrm{Recall}_j^{(t)}}}.
\end{equation}
The F1 score for one patient $j$ is by taking the average over her/his visits,
\begin{equation}
    \mathrm{F1}_j = \frac{1}{V_j}\sum_{t=1}^{V_j}\mathrm{F1}_j^{(t)}.
\end{equation}
\item \textbf{Precision Recall Area Under Curve (PRAUC)} (higher is better). To compute the PRAUC, we  treat the drug recommendation as an information retrieval problem. The PRAUC value for patient $j$ at the $t$-th visit is calculated by 
\begin{align}
    \mathrm{PRAUC}_j^{(t)} &= \sum_{k=1}^{|\mathcal{M}|}\mathrm{Precision}(k)_j^{(t)}\Delta \mathrm{Recall}(k)_j^{(t)},\notag\\
    \Delta \mathrm{Recall}(k)_j^{(t)} &= \mathrm{Recall}(k)_j^{(t)} - \mathrm{Recall}(k-1)_j^{(t)},
\end{align}
where $k$ is the rank in the sequence of the retrieved drugs, $|\mathcal{M}|$ is the total number of drugs, $\mathrm{Precision}(k)_j^{(t)}$ is the precision at cut-off $k$ in the ordered retrieval list, and $\Delta \mathrm{Recall}(k)_j^{(t)}$ is the change in recall from drug $k-1$ to $k$. The PRAUC for one patient $j$ is by taking the average over her/his visits,
\begin{equation}
    \mathrm{PRAUC}_j = \frac{1}{V_j}\sum_{t=1}^{V_j}\mathrm{PRAUC}_j^{(t)}.
\end{equation}
\end{itemize}


\smallskip
\noindent
\textbf{Implementation Details.} We follow the same setting as \cite{shang2019gamenet} and split  the dataset into training,
validation and test as $\frac23 : \frac16 : \frac16$ randomly.
For embedding tables, $\mathbf{E}_d$, $\mathbf{E}_p$, $\mathbf{E}_a$, we use
$dim=64$ as embedding size. For $\mbox{RNN}_d$ and $\mbox{RNN}_q$, we use gated recurrent unit
(GRU) with $64$ hidden units, since GRU requires fewer parameters but is reported to have
similar power as long short-term memory (LSTM). The dropout rate between two GRU
cells is $0.5$. For MPNN embedding module, the convolution radius $L$ is $2$. For $\mbox{NN}_1(\cdot)$ and $\mbox{MESSAGE}_l(\cdot),~\forall ~l$, we implement as one linear layer
plus basic ReLU activation, for $\mbox{NN}_i(\cdot),~i=2,3,4$, we use linear
functions ($\mbox{NN}_4$ is without bias term), for $\sigma_i(\cdot),~i=1,2,3$, they are all {Sigmoid} activation functions, and we use mean operator for $\mbox{UPDATE}_l,~\forall ~l$. We choose the hyperparameters of \method based on the validation set, where threshold
$\delta$ is $0.5$, the weight $\alpha=0.95$,
$K_p=0.05$, and acceptance rate $\gamma$ is selected $0.06$. Models are implemented in \textit{PyTorch 1.4.0} and parameters are
trained on Adam optimizer \cite{kingma2014adam} with learning rate $2\times 10^{-4}$ for 50 epochs. We use
a Linux machine with configuration: 30GB memory, 12 CPUs and one 16GB NVIDIA Tesla V100 GPU to implement the experiment.

In the evaluation, instead of conducting cross-validation, bootstrapping sampling is applied. First, all models are trained on a fixed training set and hyperparameters are selected from a fixed validation set. Then, with replacement, we randomly sample 80\% data points from the test set as one round of evaluation. We conduct 10 sampling-evaluation rounds and report the mean and standard deviation values.

\subsection{Additional Experiment 1: Error Analysis} 

Analyzing where and why the model
works or fails is an important step towards deep understandings. In this
section, we perform model analysis on DDI rates and number of medications/drugs in
terms of the test data. Our conclusion is that \method performs better in the scenario when a visit has lower actual DDI rate or involves more medications.

\begin{figure*}[t!] \centering
	\includegraphics[width=6.8in]{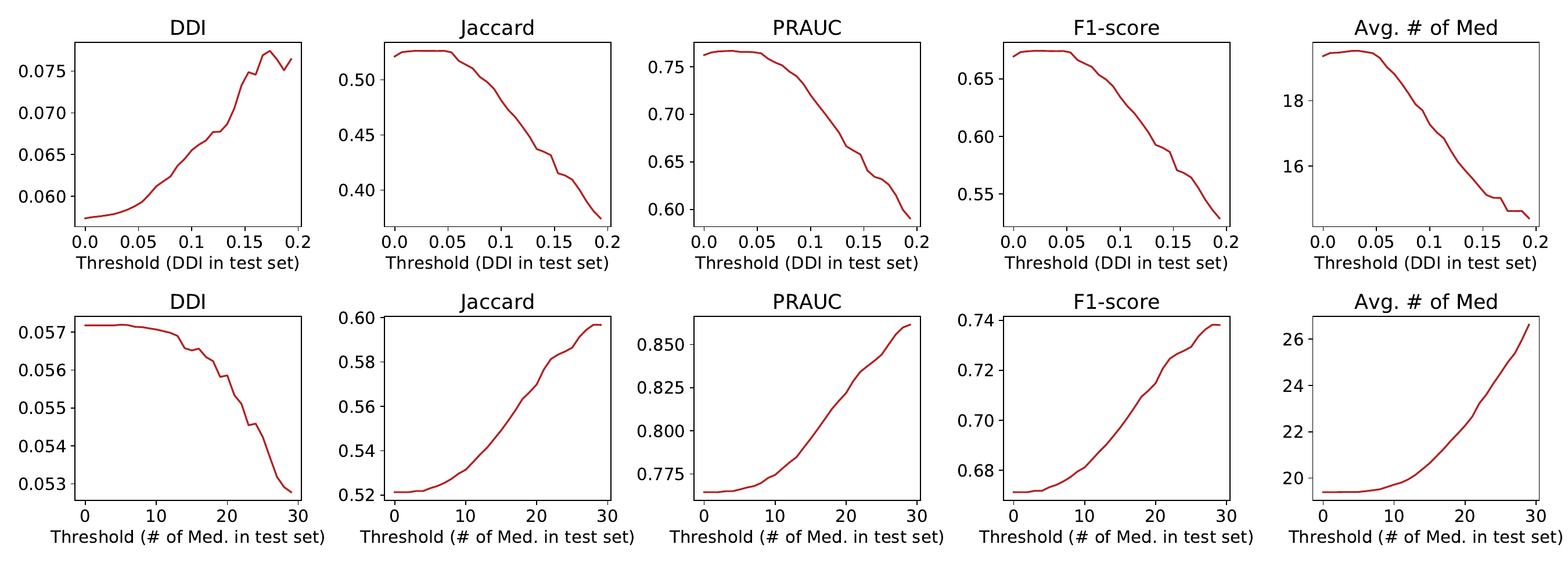}
	\caption{Performance Analysis with respect to DDI and \# of Med. in Various
		Test Scenarios 
		} \label{fig:analysis} \end{figure*}

Formally, we  evaluate on the following two
scenarios based on the whole test set: (i) we use ground truth DDI ($0.00\sim0.20$) as threshold and
remove those test points with lower DDI rates; (ii) use \# of medications
($0\sim30$) as threshold and remove those test points with smaller drug
set. For each test scenario, we calculate four evaluation metrics plus the avg. \# of med. and draw the trends in Figure~\ref{fig:analysis}.

When the DDI threshold increases or the
medication threshold decreases, the overall performance will drop. The reason
might be that on the one hand, if the ground truth DDI is high, our model tends
to recommend fewer drugs for a lower DDI combination, thus leading to a
sacrifice on the accuracy; on the other hand, when formulating the task as a
multi-label prediction problem, a smaller ``\# of med" threshold means less data and more noise, which undermines the model.

\subsection{Additional Experiment 2: Effectiveness of Mask Matrix to avoid Interacted Drugs} \label{sec:case1}

In this section, we hope to interpret the benefit of bipartite mask $\mathbf{H}$. For preparation, we load the pre-trained \method model and extract the parameter matrix, $\mathbf{W}_4\odot\mathbf{H}$, denoted by $\mathbf{W}\in\mathbb{R}^{|\mathcal{S}|\times|\mathcal{M}|}$. We inspect the {\em learned weights} correspond to interacted drug pairs. From experiments, we will see the interacted drugs have negative-correlated weights, which means {\em they are less likely to be co-prescribed by our model.} Below are more details.

\smallskip
\noindent
\textbf{Basics of $~\mathbf{W}$.} On average, when constructing the bipartite structure $\mathbf{H}$ by fragmentation, each drug (right cells in the bipartite architecture in Figure~\ref{fig:framework}) is connected to 6.84 out of $|\mathcal{S}|=$ 491 functional substructures (left cells) and conversely, each substructure is connected to 1.82 out of $|\mathcal{M}|=$ 131 drugs. Thus, $\mathbf{W}$ is a sparse matrix. Each element of $\mathbf{W}$, for example, $\mathbf{W}_{ij}$ quantifies the weight relation between a  substructure $i$ and a particular drug $j$.

\smallskip
\noindent
\textbf{Evaluation Procedure.} For drug $j$, its functional weight representation is the $j$-th column of the weight matrix, $\mathbf{W}_{:j}$, a {\em sparse} real-valued vectors of size, $|\mathcal{S}|$, where $\mathcal{S}$ is the overall substructure set.
To evaluate the efficacy of DDI reduction, we hope that for a pair of interacted drugs, e.g., $i$ and $j$, the cosine similarity of their weight representation is negative, i.e., 
\begin{equation}
    cos(\mathbf{W}_{:,i}, \mathbf{W}_{:,j}) = \frac{\mathbf{W}_{:,i}\cdot\mathbf{W}_{:,j}}{\|\mathbf{W}_{:,i}\|\cdot\|\mathbf{W}_{:,j}\|}<0,
\end{equation} 
which prevents their co-prescription. For {\em all  pairs of interacted drugs}, we design the average cosine similarity, $Cos_{interacted}$, 
\begin{equation}
    Cos_{interacted} = \frac{\sum_{i,j:~\mathbf{D}_{ij}=1} cos(\mathbf{W}_{:i}, \mathbf{W}_{:j})}{\sum_{i,j:~\mathbf{D}_{ij}=1}1},
\end{equation}
where $\mathbf{D}$ is the DDI matrix. And for {\em all possible drug pairs}, we design the average cosine similarity, $Cos_{all}$,
\begin{equation}
    Cos_{all} = \frac{\sum_{i,j:~i\neq j} cos(\mathbf{W}_{:i}, \mathbf{W}_{:j})}{\sum_{i,j:~i\neq j}1}.
\end{equation}
Following the intuition, we hope that $Cos_{interacted}$ could be much smaller than $Cos_{all}$. To conduct ablation study for mask $\mathbf{H}$, we remove it and run the experiment again with the settings reported in Section~\ref{sec:setting}. We show the $Cos_{interacted}$ and $Cos_{all}$ as well as the output DDI of two cases in Table~\ref{tb:competing}. 
		
\begin{table}[h!] \centering \caption{$Cos_{interacted}$, $Cos_{all}$ and output DDI vs Mask $\mathbf{H}$} 
		\begin{tabular}{c|ccc} 
		\toprule 
		with mask $\mathbf{H}$ & $Cos_{interacted}$ & $Cos_{all}$ & output DDI\\ 
		\midrule 
		
		Yes & -0.0030 & 0.0021 & 0.0589 \\
		No & 0.0036 & 0.0017 & 0.0652\\
		\bottomrule 
		\multicolumn{4}{c}{
          \begin{minipage}{7cm}
            \vspace{1mm}
            \noindent~* With mask $\mathbf{H}$, $Cos_{interacted}<0<Cos_{all}$ is observed, which indicates that the interacted drugs are less likely to be co-prescribed. Without mask $\mathbf{H}$, the model cannot identify and prevent interacted drugs. In practice, the case without mask $\mathbf{H}$ may also have low DDI because of our proposed controllable loss function.
          \end{minipage}
        }
        \end{tabular}
	\label{tb:competing}\end{table}
	
\smallskip
\noindent
\textbf{Results and Interpretation.} As is mentioned, $\mathbf{W}$ is sparse, so the calculated cosine similarities all have low magnitudes, and it does not necessarily mean that the correlation is not significant. It is interesting that \method with mask $\mathbf{H}$ have $Cos_{interacted}<0<Cos_{all}$, which means the mask does enable the model to send negatively-correlated signals for interacted drugs, so that the output DDIs decrease in the final combination. We  observe that the model without $\mathbf{H}$ (i.e., $\mbox{NN}_4$ is a fully connect network) 
cannot distinguish interacted drug pairs, and the correlation value $Cos_{interacted}$ of the interacted drugs is even larger than $Cos_{all}$ of all drug pairs, which is undesirable and counter-intuitive. The reason why its DDI value is also lower than other baselines is due to the DDI loss and our proposed controllable loss function.

\end{document}